# The Fifteen Puzzle- A New Approach through Hybridizing Three Heuristics Methods

**Dler O. Hasan** [1,] https://orcid.org/0000-0001-9457-9655**, Aso M. Aladdin** [1,2] https://orcid.org/0000-0002-8734-0811**, Hardi Sabah Talabani** [1] https://orcid.org/0000-0001-6042-0064, **Tarik Ahmed Rashid** [3*] https://orcid.org/0000-0002-8661-258X, **Seyedali Mirjalili** [4,5] https://orcid.org/0000-0002-1443-9458

1 Department of Applied Computer, College of Medical and Applied Sciences, Charmo University, Sulaymaniyah, KR, Iraq. dler.osman@charmouniversity.org; aso.aladdin@charmouniversity.org; hardi.sabah@charmouniversity.org.
2 Department of Technical Information Systems Engineering, Erbil Technical Engineering College, Erbil Polytechnic University, Erbil, KR, Iraq. aso.dei20@epu.edu.iq
3 Computer Science and Engineering Department, University of Kurdistan Hewler, Erbil, KR, Iraq. tarik.ahmed@ukh.edu.krd
4 Centre for Artificial Intelligence Research and Optimization, Torrens University, Adelaide, Australia.
5 Yonsei Frontier Lab, Yonsei University, Seoul, Korea.
ali.mirjalili@gmail.com

* Correspondence: tarik.ahmed@ukh.edu.krd;

**Abstract:** Fifteen Puzzle problem is one of the most classical problems that have captivated mathematical enthusiasts for centuries. This is mainly because of the huge size of the state space with approximately $10^{13}$ states that have to be explored and several algorithms have been applied to solve the Fifteen Puzzle instances. In this paper, to deal with this large state space, Bidirectional A* (BA*) search algorithm with three heuristics, such as Manhattan distance (MD), linear conflict (LC), and walking distance (WD) has been used to solve the Fifteen Puzzle problems. The three mentioned heuristics will be hybridized in a way that can dramatically reduce the number of generated states by the algorithm. Moreover, all those heuristics require only 25KB of storage but help the algorithm effectively reduce the number of generated states and expand fewer nodes. Our implementation of BA* search can significantly reduce the space complexity, and guarantee either optimal or near-optimal solutions.

## 1 Introduction

The Fifteen Puzzle is a standard sliding puzzle invented by Samuel Loyd in the 1870s [1], which consists of 15 tiles with one tile missing within a 4x4 grid. The fifteen tiles numbered from 1 to 15. The numbered tiles should be initially ordered randomly. This game aims to slide those tiles, which are located next to the space into the space (one at a time) to get the numerical order of the tiles from left to right with the blank at the bottom right/top left corner in a minimum time and moves. Automatically solving the Fifteen Puzzle is very challenging because the state space for the Fifteen Puzzle contains about $16!/2 \approx 10^{13}$ states [2]. 15-puzzle contains 16! instances but only half of the instances are solvable [3], [4]. Optimal solutions for any solvable instances of the Fifteen Puzzle can take from 0 to 80 moves [5] [6]. The two common heuristic search algorithms, such as A* [7] and Iterative Deepening A* (IDA*) [8] have been successfully used for computing optimal solutions for the Fifteen Puzzle Instances. Those algorithms are guided by heuristic functions, which are estimates of the number of moves required to solve any given puzzle configuration.

The most common heuristic functions that have been used to reduce the search space are misplaced tile (MT), MD, LC, and Pattern Databases (PDBs) [9],[10],[11], and WD has also been used but this one is not common. Misplaced tile (MT) is the number of tiles that are not in their goal positions. MD is the sum of the distance of each tile from its goal position. LC is the sum of two moves for each pair of conflicting tiles, which are in their goal row or column positions but in the wrong order. WD was developed by [12], which counts the vertical moves and horizontal moves separately while considering the tiles' conflict with each other. PDBs are heuristics in the form of lookup tables. The two heuristics misplaced tile and Manhattan distance were used with the A* algorithm for optimally solving the 8 puzzle problems by [13]. The Manhattan distance and linear conflict heuristics were combined and used with IDA* algorithm for the Fifteen Puzzle by [14]. The walking distance heuristic was developed and used (with IDA* search) by [12] for Fifteen Puzzle. To the best of our knowledge, the walking distance heuristic has not been used in any research for Fifteen Puzzle. Pattern database heuristics were firstly introduced by [15] and then used by many researchers, and now there are various types of pattern databases [16]. The main drawback of pattern databases is that they require a large amount of memory (several gigabytes for some types of pattern databases) [16], [17]. Flener, Korf, and Hanan [18] claimed that an effective heuristic for the Fifteen Puzzle is the 7-8 additive pattern database, but this heuristic requires a lot of storage space and can be memory intensive which is about 575 megabytes.

All the heuristics used to estimate how close a state is to the goal suffer from several drawbacks. For example, some of them are not very accurate to estimate the remaining distance to a goal such as MD, MT, and WD, and the others are accurate but they require a lot of storage space such as PDBs. The main objective of this paper is to combine some heuristics to accurately estimate the cost from the current state to the goal state without generating a lot of states or requiring a large amount of storage space to store the nodes. The contribution of this paper is to hybridize three heuristics MD, LC, and WD to estimate the number of steps to the goal state. Moreover, to increase the effectiveness of the heuristic function, the MD value is divided by three. We use this heuristic in such a way as to significantly reduce the number of generated nodes to solve the puzzle states. Using those heuristic algorithms in that way cannot be guaranteed to give an optimal solution, but they usually find an optimal solution or a solution that is 1 to 6 moves far from the optimum and in some rare cases more than 6 moves away from the optimum. For the most difficult states, we run two searches – a forward search from the initial state and a backward search from the ending state (goal state), which is called a bidirectional search. This is for the sake of improving the algorithm's performance.

The remainder of this paper is structured as follows. Section 2 is devoted to presenting and discussing the implementation of our BA* algorithm. Section 3 presents and evaluates the three heuristics we use to solve the Fifteen Puzzle problem. Section 4 presents the efficient way of hybridizing the three heuristics for solving the Fifteen Puzzle. Section 5 presents and discusses the results and their comparisons. Section 5.1 compares our implementation of the BA* algorithm with the Artificial Bee Colony (ABC) algorithm in terms of efficiency and inadmissibility. Section 5.2 discusses the comparison between Bidirectional A* (BA*) search and Unidirectional A* (UA*) search. Section 5.3 describes the experiments performed with our implementation of the BA* algorithm and it also compares the obtained results of our algorithm with the results obtained by IDA* algorithm with MD and LC heuristics. Finally, Section 6 highlights the main conclusion of this study

## 2 Bidirectional A* Algorithm

IDA* and A* are the two most popular heuristic search algorithms widely used to solve the Fifteen Puzzle problems. A* algorithm is one of the most well-regarded algorithms in artificial intelligence for finding the shortest path or the smallest number of moves from the initial state to the goal [7]. Despite being complete, this algorithm has some disadvantages that can make that algorithm inefficient, especially for complex and large puzzle problems. This is because for the difficult states billions of nodes need to be expanded and generated, and in the A* algorithm all the generated nodes are kept in memory, which can lead to running out of memory or sometimes finding a solution takes a long time. IDA* algorithm is a variant of the A* algorithm that can be implemented for solving Fifteen Puzzle [8]. Due to the reason that IDA* does not store the expanded nodes in memory, it uses less space and expands nodes faster than the A* algorithm. Even though IDA* algorithm is more efficient than the A* algorithm, we still use the A* algorithm in this paper for some reason. First of all, since we use bidirectional search, the A* algorithm is a good choice because it stores all the generated nodes in memory and this leads to frontier intersections that can be easily tested [9]. Second of all, the A* algorithm with those heuristics that we use generates a few states, and this does not cause the algorithm to run out of memory. Thirdly, since the A* algorithm retains the generated states in memory, each state is generated once. Algorithm 1 gives the pseudocode for Bidirectional A* (BA*) algorithm. Some notations are used such as OpenList, ClosedList and NeighboringState denote the states that have been visited but not expanded, the states that have been visited and expanded, and the state that is directly connected to the current state. There are separate copies of those variables for both forward and backward search, with a subscript (F or B) indicating the direction:

Forward search: $\text{OpenList}_f$, $\text{ClosedList}_f$ and $\text{NeighboringState}_f$, etc.

Backward search: $\text{OpenList}_b$, $\text{ClosedList}_b$ and $\text{NeighboringState}_b$, etc.

BA* algorithm for each one of the two searches (forward and backward search) needs two lists: a closed list which is used for storing all the puzzle states that have been visited and expanded and an open list which is used for storing the puzzle states that have been visited but not expanded. At each step, the heuristic value and the depth cost of the current state is determined. Then, the states inside the open list are sorted according to the heuristics value in increasing order. At every step, the head of the open list which has the lowest evaluation function value (which is the heuristic value plus the path cost) is removed from the open list and then checked whether it is the goal state (start state for backward search) or not. If the head state is the goal state (start state for backward search), the algorithm reconstructs the path to the goal (to the start for backward search). If the head state is not the goal (is not the start for backward search), it is checked if it is in the closed list of the opposite search direction, and if there is, it reconstructs the solution path from the two searches. When the goal (start for backward search) was not found, the head state is expanded (all the valid moves are specified) and on the closed list, it is placed. Then, all the successors of the head state which are not already on the closed list are stored in the open list. As it is shown in Algorithm 1, the forward search starts first and continues until 75,000 states are expanded but after the first step of the cycle, the forward search continues until 15,000 states are expanded. If the solution path from the start state to the goal state was not found during generating those number of states, the forward search stops, and the backward search starts. The backward search continues until 75,000 states are expanded (until 15,000 states are expanded after the first step of the cycle) if, during that period the solution path from the goal state to the start state was not found, the backward search stops, and the forward search starts again. This process will continue until the solution path is found.

**Algorithm 1 BA* algorithm pseudocode**

**function** BA*(*StartState*, *GoalState*)
  **Initialise:**
    $Iterator_f$ to control the loop
    $OpenList_f$ to store the states to be traversed
    $ClosedList_f$ to store already traversed states
    $OpenList_b$ to store the states to be traversed
    $ClosedList_b$ to store already traversed states
  **if** $Iterator_f$ = 0 **then**
    set depth cost of *StartState* (g(s) in Equation (2)) to zero
    calculate HH value from *StartState* to *GoalState*. Equation (3)
    calculate evaluation function for *StartState*. Equation (2)
    add *StartState* into $OpenList_f$ and $ClosedList_f$
  **while** $OpenList_f$ is not empty **do**
    $CurrentState_f$ is state with lowest evaluation function value (Equation (2)) in $OpenList_f$
    remove $CurrentState_f$ from $OpenList_f$
    **if** $CurrentState_f$ is *GoalState* **then**
      reconstruct the solution path from *StartState* to $CurrentState_f$, and terminates the loop
    **for each** $NeighboringState_f$ of $CurrentState_f$ **do**
      **if** $NeighboringState_f$ is not in $ClosedList_f$ **then**
        depth cost of $NeighboringState_f$ is equal to the depth cost of $CurrentState_f$ plus one
        calculate HH value from $NeighboringState_f$ to *GoalState*. Equation (3)
        calculate evaluation function for $NeighboringState_f$. Equation (2)
        add $NeighboringState_f$ into $ClosedList_f$
        add $NeighboringState_f$ into $OpenList_f$
        **if** $NeighboringState_f$ is in $ClosedList_b$ **then**
          reconstruct the solution path from the two searches: from *StartState* to $NeighboringState_f$ and form $NeighboringState_f$ to *GoalState*, and terminates the loop
        increase $Iterator_f$ by 1
    **if** $Iterator_f$ mod 15000 is equal to 0 after the first step of the cycle or $Iterator_f$ mod 75000 is equal to 0 **then**
      ->Expand in the backward direction, analogously

## 3 Heuristic Functions

A heuristic is an informed guess to choose the next node to visit when exploring a search space. A heuristic can lead the algorithm to a solution or fail to reach the goal. The three heuristics which are used in this paper are Manhattan distance, walking distance, and linear conflict. Figure 1: Panel (a) shows an arbitrary start state of the Fifteen Puzzle and panel (b) shows the goal state of the Fifteen Puzzle. The tiles are denoted by $t_i$ and the blank by $t_0$. $<t_1, t_4, t_2, t_3, t_{13}, t_6, t_7, t_8, t_5, t_{10}, t_{11}, t_0, t_9, t_{14}, t_{15}, t_{12}>$ for start state and $<t_1, t_2, t_3, t_4, t_5, t_6, t_7, t_8, t_9, t_{10}, t_{11}, t_{12}, t_{13}, t_{14}, t_{15}, t_0>$ for goal state shown in the Figure 1.

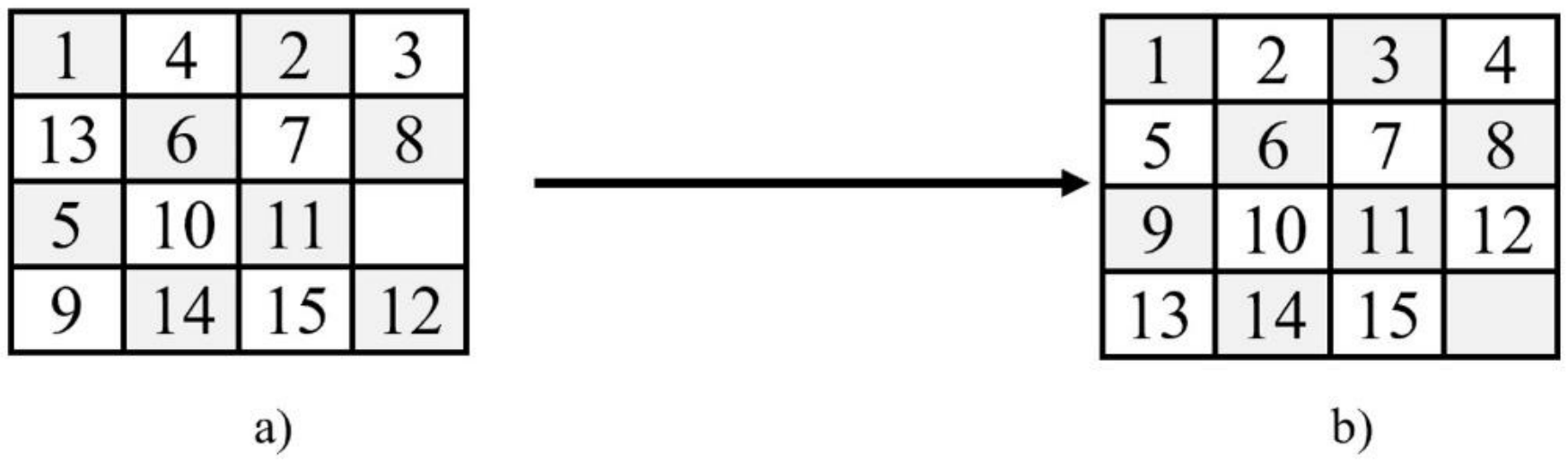

**Figure 1. Fifteen Puzzle a) start state, b) goal state**

The Manhattan distance of a puzzle is the sum of the horizontal and vertical distance of each tile (except the blank tile) from its goal position [8]. For the initial state of the Fifteen Puzzle shown in Figure 1, only the tiles $t_4$, $t_2$, $t_3$, $t_{13}$, $t_5$, $t_9$, and $t_{12}$ are not in their goal positions, and they are away from their goal positions by 2, 1, 1, 2, 1, 1 and 1 respectively. Therefore, the heuristic function evaluates to 9 (2+1+1+2+1+1+1). This means that the current state needs at least 9 moves to reach the goal. Manhattan distance is admissible because it never overestimates the number of moves to the goal and each tile must at least be moved from its current position to its goal position, and only vertical and horizontal movement is allowed. Therefore, the Manhattan distance value of any state is less than or equal to the number of moves that the state needs to reach the goal. The Manhattan distance of a tile in a puzzle can be found using Equation (**1**) ($s$ is the current state) [17]:

$$h(s) = \sum_{i=1}^{n} (|x_i(s) - \overline{x_i}| + |y_i(s) - \overline{y_i}|) \tag{1}$$

LC which is used to enhance the effectiveness of the Manhattan distance, adds two additional moves to the Manhattan distance for each pair of conflicting tiles that would have to be swapped to reach the goal state. Two tiles $t_i$ and tj are in a linear conflict if both tiles are positioned in their goal row or column but in the wrong order or other words, they are reversed relative to their goal location [14]. For example, in Figure 1, tile $t_4$ conflicts with tiles $t_2$ and $t_3$ because by changing the row of tile $t_4$ we can eliminate all conflicts, and tile $t_{13}$ conflicts with tiles $t_5$ and $t_9$ because they are in the correct column but in inverse order. In that case, $t_9$ must do one move right to let the others pass by and then back to its column position. These four moves are not counted in Manhattan distance. Therefore, two additional moves are added to the Manhattan Distance for each pair of conflicting tiles and the heuristic evaluation function remains admissible.

Up until now the total cost function for the initial state in Figure 1 is equal to 13 (9 for Manhattan distance, 4 for linear conflict) while the optimal solution for the initial state is 29 moves. Therefore, using those two heuristics cannot make the algorithm efficient, especially for complex and large puzzle problems, and finding the solution takes a long time. This is because Manhattan distance does not capture the conflictions and interactions between the tiles, and this leads to heavily underestimating the actual optimal solution cost in almost all the problem instances of the Fifteen Puzzle figure [19], and linear conflict only adds two moves for every two tiles which are positioned in the correct row/column, but inverted. The walking distance counts the vertical moves and horizontal moves separately while considering the tile's conflict with each other [12]. According to the goal state in Figure 1, on the first row of the initial state, all the 4 tiles ($t_1$, $t_4$, $t_2$, and $t_3$) are from the 1st row of the goal state, and 0 tiles from the other rows of the goal state. The same approach is used for the other rows as is shown in Table 1.

**Table 1. Walking Distance calculation**

| **No. of rows** | **Number of tiles from 1st row** | **Number of tiles from 2nd row** | **Number of tiles from 3rd row** | **Number of tiles from 4th row** | **Blank tile** |
|---|---|---|---|---|---|
| **1st row** | 4 | 0 | 0 | 0 | |
| **2nd row** | 0 | 3 | 0 | 1 | |
| **3rd row** | 0 | 1 | 2 | 0 | ← here |
| **4th row** | 0 | 0 | 2 | 2 | |

To calculate the horizontal walking distance, we can only swap the blank tile with any single tile from the above or below row and the order of the tiles on each row is irrelevant. We keep swapping until all the tiles are in their goal rows. The minimum number of moves needed to take all the tiles to their goal row positions is the horizontal Walking Distance. We can apply the same procedure to calculate the vertical Walking Distance by taking all the tiles to their goal column positions with the minimum number of moves and each tile can only be taken in a column adjacent to the column containing the blank tile, and swapping places with it. The order of the tiles on each column is irrelevant. The total walking distance is the sum of the number of horizontal and vertical moves. To more explain Table 1, Figure 2 illustrates how WD can be calculated manually step by step for the initial state in Figure 1. Two 4 by 4 tables are needed one for computing horizontal WD value and another one for computing vertical WD value.

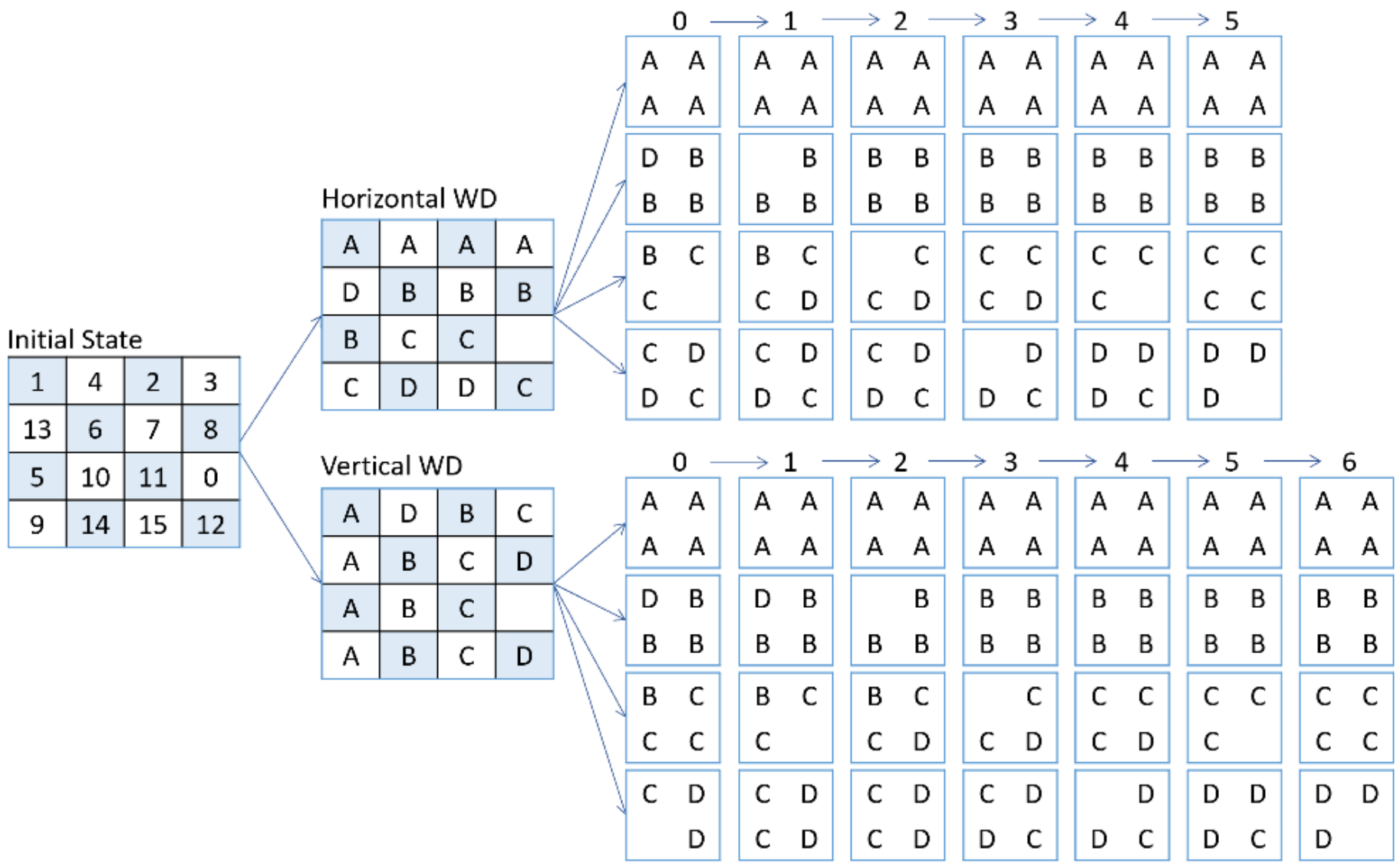

**Figure 2. Step by step Walking distance calculation**

The table of the horizontal WD in Figure 2 has 4 'A' elements on the first row, this means all the tiles $t_1$, $t_4$, $t_2$, $t_3$ are from the 1st row of the goal state. It has 1 'D' element with 3 'B' elements on the second row, this is because $t_{13}$ is from the 4th row of the goal, and $t_6$, $t_7$, $t_8$ are from the 2nd row of the goal. It has 1 'B' element, and 2 'C' elements with a blank ($t_0$) element on the third row, this is because $t_5$ is from the second row of the goal, and $t_{10}$, $t_{11}$ are from the third row of the goal. It has 2 'C' elements with 2 'D' elements on the fourth row, this is because $t_9$, $t_{12}$ are from the third row of the goal, and $t_{14}$, $t_{15}$ are from the fourth row of the goal. As is shown in Figure 2, only five steps are needed to take all the tiles to their goal row position, this is the value of horizontal WD value for the initial state in Figure 1. The same procedure is used for building the table of vertical WD and calculating its vertical WD value except that when we build the table of vertical WD we must specify each tile from which column of its goal position is. As it can be seen in Figure 2, to take all the tiles to their column position six steps are needed, this is the value of vertical WD value for the initial state in Figure 1. The total walking distance is the sum of the number of horizontal and vertical moves which is 11 steps.

Since walking distance cannot be easily computed at runtime, we can precompute all these values and store them in the database because if we do not pre-compute them, this heuristic can slow the search down significantly. Instead of full calculation of walking distance during the search, Breadth-First Search (BFS) can be executed backward from the goal state to obtain all the distinct tables for all the Fifteen Puzzle configurations (all possible configurations of the tiles) which are only 24964 patterns, and store them in the database to speed up the search. The size of the database is relatively small which is about 25KB. The same database is used for calculating the number of horizontal and vertical moves. The maximum walking distance value is 70 (such as $t_0$, $t_{15}$, $t_{14}$, $t_{13}$, $t_{12}$, $t_{11}$, $t_{10}$, $t_9$, $t_8$, $t_7$, $t_6$, $t_5$, $t_4$, $t_3$, $t_2$, $t_1$), 35 moves for each horizontal and vertical moves. WD is more accurate and efficient than the Manhattan distance because the WD value is always greater than the MD value as it is illustrated in Figure 3 (The data in Figure 3 can be seen in Table 6).

Figure 3 shows MD, WD, and Optimal values for Korf's 100 instances [8] after sorting the instances by optimal value. In all of them, the WD value was greater than the MD value and Table 2 shows that the total WD value for all the 100 instances is greater than the total MD value. Table 2 also shows the minimal total cost (optimal solution) and total LC values for all 100 instances.

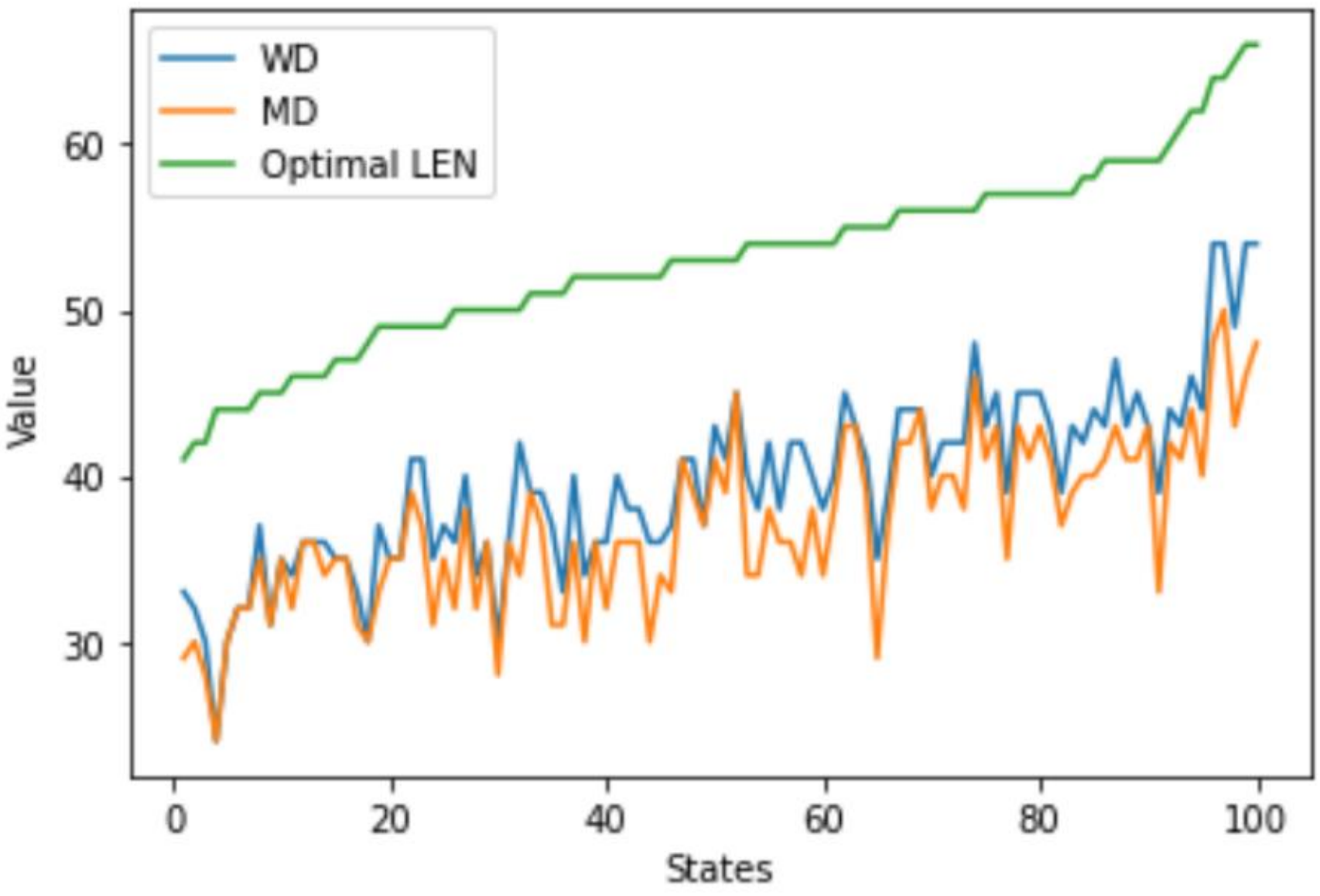

**Figure 3. MD, WD, and optimal value for Korf's 100 instances**

**Table 2. Total WD, MD, LC, and Optimal solution lengths for Korf's 100 instances of Fifteen Puzzle**

| Problems | Total WD | Total MD | Total LC | Total Optimal |
|---|---|---|---|---|
| **Korf's 100 instances of 15-Puzzle** | 3957 | 3705 | 188 | 5307 |

The walking distance can also be enhanced by linear conflict because WD does not count the two moves which are determined by Linear conflict for each pair of conflicting tiles. As shown in Figure 2 during calculating the horizontal or vertical WD values when we have two tiles, which are in linear conflict, the first tile can slide to the above or below row if the row contains a blank without removing the second tile, and for the second tile is also correct. For example, Figure 4 zooms in and shows a part of Figure 2 where the tile $t_{13}$ (D) that conflicts with the tile $t_5$ (B) can slide to the third row without removing the tile $t_5$ (B).

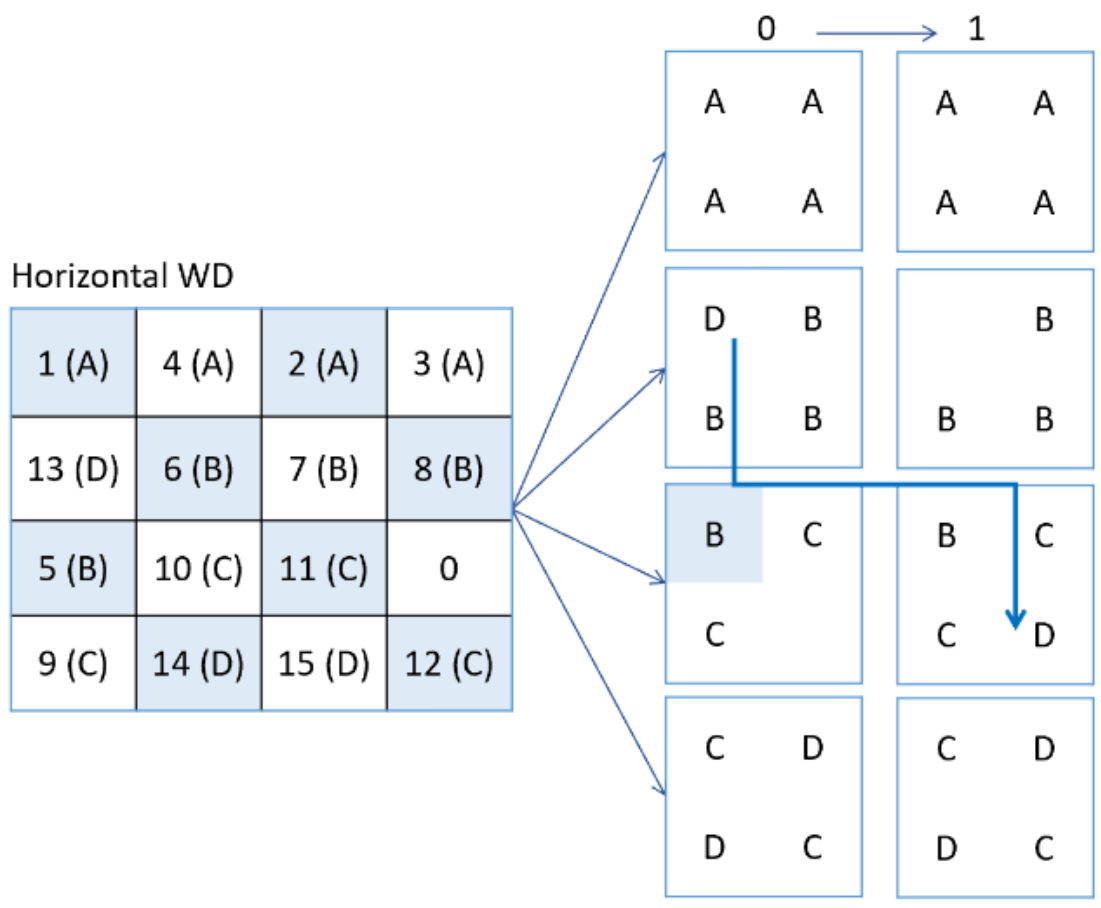

**Figure 4. WD does not capture LC heuristic**

We have built the Walking Distance lookup table for both goal states with a blank at the bottom right and top left corner since the two different goal states have been used in many types of research and we also use those two different goal states in this paper.

# 4 Hybridized Heuristic Functions

Since we have no perfect heuristic function (exact distance function) to give us the exact number of moves needed to solve all the Fifteen Puzzle instances and each heuristic has its way to calculate the distance between the current state to the goal state, it is desirable to combine multiple heuristics which can complete each other to estimate the solution cost. To more accurately estimate the cost of reaching the goal, combining several heuristics is generally the best way but it is challenging [20], [21], [22]. Multi-heuristic has been used in different ways. The most common way to use multiple heuristics is to combine different heuristics and use their maximum value. Holte et al [16] showed that taking the maximum heuristic value among several heuristics can lead to reduce node generation and result in improving the performance of the search. When two or more admissible heuristics are combined, taking their maximum, that is, by defining ($\mathrm{hmax}(s) = \max(\mathrm{h1}(s), \mathrm{h2}(s))$) is also admissible [16], [23]

Another way to use multiple heuristics is cost partitioning or cost splitting, which has been used by many researchers [24], [25], [26] which is a technique to add multiple heuristic values in an admissible way with operator cost partitioning, by distributing the cost of each operator among them. This technique has a drawback in finding good cost partitioning [25]. Korf and Taylor [27] took advantage of several heuristics including Manhattan distance, linear conflict, last moves, and corner-tile to improve the accuracy of the heuristic evaluation function and result in improving the search performance of the IDA* search. In addition, they used the heuristics in a way that keeps the heuristics still admissible for example, when the same tile is involved in a corner tile and linear conflict, the extra moves are added only one time. Therefore, whenever we combine multiple heuristics and we want to find the optimal solution, we must be sure that an actual distance for any tile is not calculated more than one time. Those heuristics are not complex and it can be easily checked what tiles are involved in multiple heuristics. Manhattan distance and walking distance are the two heuristics that we use in this paper are complex and it is not easy to check which tile's actual distance to its goal position is counted by the two heuristics.

Each heuristic has its strength and weakness. Therefore, we must determine the weakness and strengths of the heuristics when we want to combine multiple heuristics to create a more accurate heuristic function. The main drawback of Manhattan distance is measuring each tile's distance to its goal position without considering interference from any other tiles [18]. For example, according to MD, the tiles $t_6$, $t_7$, $t_8$, $t_{10}$, $t_{11}$, $t_{14}$, and $t_{15}$ in the initial state shown in Figure 2 need zero moves to reach their actual positions since they are already in their goal positions. This estimation is not correct because it is not possible to take the tiles $t_4$, $t_2$, $t_3$ $t_{13}$, $t_5$, and $t_9$ to their goal positions without moving some of these tiles $t_6$, $t_7$, $t_8$, $t_{10}$, $t_{11}$, $t_{14}$, and $t_{15}$. On the other hand, WD considers interactions between tiles, and in some way, it calculates the distance of the tiles to their goal positions like MD. As is illustrated in Figure 5 which is a part of Figure 2, one of these tiles $t_7$, $t_{11}$, and $t_{15}$ makes two moves while calculating the WD value, this proves that WD is more efficient than MD.

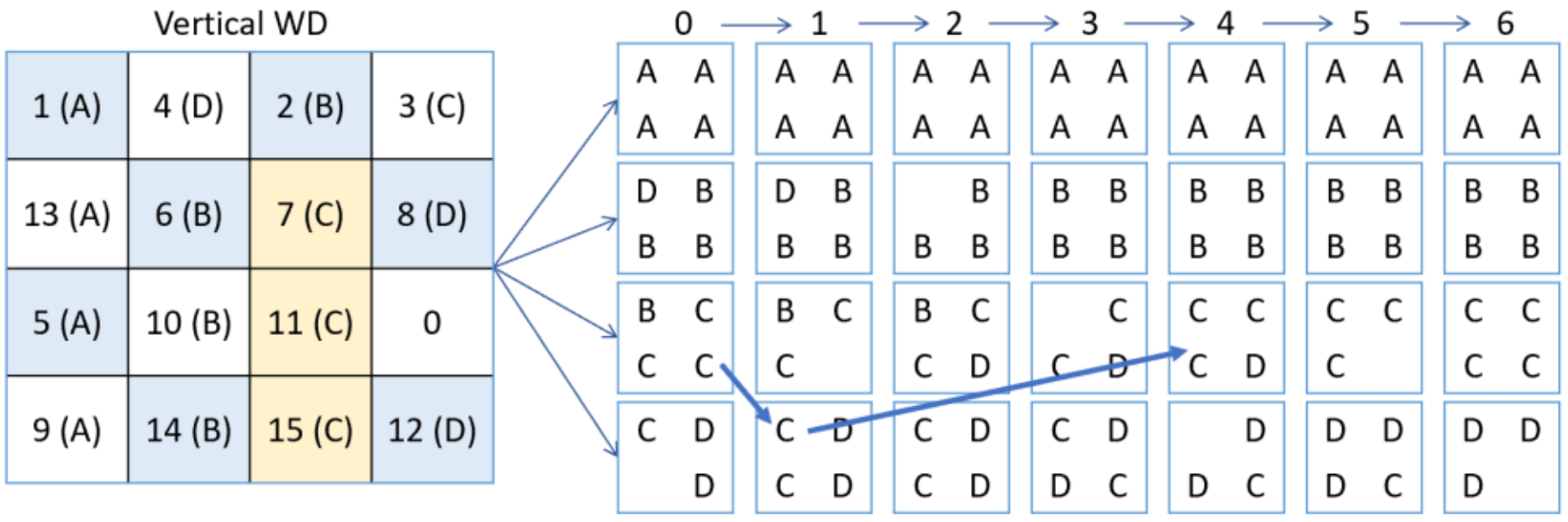

**Figure 5. MD considers the interference of tiles with each other**

WD is not exactly equal to MD plus the interference of tiles with each other. It seems that the WD heuristic considers interactions between tiles and the distance of each tile to its goal position but we think it cares more about the interaction of tiles than their distances to their end positions because there are many Fifteen Puzzle problem instances that have the same MD and WD value. For instance, there are 23 instances in Table 6 that have the same WD and MD values despite a lot of interactions between their tiles. Additionally, for this instance $<t_{15}, t_4, t_7, t_{11}, t_5, t_8, t_0, t_3, t_{14}, t_2, t_{12}, t_{13}, t_1, t_6, t_{10}, t_9>$, the WD and MD are 35 while there are a lot of interactions between the tiles. This proves that if the WD is equal to the exact MD plus the conflicts between the tiles, the WD value for the Fifteen Puzzle instance must be greater than but not equal to the MD value. Therefore, the WD is not equal to the exact MD plus the conflicts between the tiles for the Fifteen Puzzle problem instances. Despite the previous reason, WD works somehow but not exactly as MD since WD takes each tile to its goal column-row position when calculating the horizontal and vertical values for a puzzle instance as it is illustrated in Figure 2. In general, WD is more efficient and better than MD because WD is never less than MD as it is illustrated in Figure 3 and Table 6. Because of that reason we use WD and LC as the main heuristics together with MD as a helping heuristic to assist the main heuristics. Since we use MD as a helping heuristic, the MD value is divided by 3 and in that way the MD value is reduced to a number when it is added to the main heuristics' value, the result will be close to the optimal solution length.

As it was explained before WD mainly considers the interactions between the tiles and in some way it calculates the distance of the tiles to their actual positions like MD. Therefore, to compensate for the calculating tiles' distance to their goal position MD is used but not the whole MD value. The MD value is divided by a number (which is three) so that the summation of WD, LC, and MD/3 will be close to the optimal solution length. For example, if we sum the total WD value (3957), LC value (188), and MD value divided by three (3705/3) as it is shown in Table 2, the result will be 5,380, and this result is very near to the total optimal solution value 5307 for the Korf's standard 100 random Fifteen Puzzle instances. Furthermore, this total overestimation is very small and it does not have a great impact on the results of BA* as Table 6 shows that 91% of the instances are 0 to 6 moves away from their optimal solutions and reaching the goal for each instance, a small number of states are generated. Because of that reason we calculate the heuristic function in the evaluation function (Equation (**2**) [1]) as shown in Equation (**3**) named as Hybridizing Heuristic (HH). To find the shortest path, the A* algorithm uses the evaluation function as it is shown in Equation (**2**) which is equal to $g(s)$ the depth cost from the start state to the current state plus the $h(s)$ the heuristic that estimates the distance from current state to the goal state. A* algorithm gurrantees optimal solution if the heuristic function is admissible.

$$f(s) = g(s) + h(s) \tag{2}$$

$$h(s) = \frac{md(s)}{3} + wd(s) + lc(s) \tag{3}$$

# 5 Results and Discussions

In this section to evaluate the efficiency and performance of our implementation of the BA* algorithm, we make some comparisons. Firstly, BA* with HH is compared with the ABC algorithm in terms of admissibility. Secondly, in terms of directionality, BA* and UA* are compared to show that bidirectional search is more efficient than unidirectional search especially when there is a guarantee that the two searches of bidirectional do not pass by each other without intersecting search and they meet. Finally, the BA* search with HH is run on Korf's 100 instances, along with the comparison with IDA* search.

## 5.1 Inadmissible Heuristics

An algorithm can guarantee to find the shortest path or the smallest number of moves from the initial state to the goal only if the heuristic function never overestimates the actual path cost from the current state to the goal state, which we call an admissible heuristic [28]. Due to the reason that finding the optimal solution for the Fifteen Puzzle is too expensive and requires searching through a very large number of paths and generating a large number of nodes [29], many types of research have been conducted to obtain near-optimal solutions instead of exact optimal solutions [30], [31]. Thayer, Dionne, and Ruml [32] state to reduce the solving time, a near-optimal solution is a practical alternative. To reduce the number of generated nodes, we have incorporated aspects from the three heuristics to create a better one and the heuristic function in the evaluation function (Equation (**2**) [1]) is calculated as shown in Equation (**3**). As shown in Equation (**3**) three heuristics are combined to estimate the cost from a given state (node) to the goal state. The value of Manhattan distance is divided by three because calculating in that way, leads to fewer nodes to be generated during the search. Because of the previous reasons, the algorithm heuristic cannot guarantee to find the goal with the smallest number of moves but this brings some advantages. Firstly, a lesser number of nodes are generated and it can very quickly find the goal. Secondly, the result is very close to the optimal solution. Since a bidirectional search has been used to find the path from the initial state to the goal state, the three heuristics have been used in either direction (search).

However, our implementation of the BA* algorithm with the three heuristics does not find optimal solutions for most of the Fifteen Puzzle instances, the difference between the solution length found by BA* and the optimal solution for each puzzle instance does not increase when the puzzle instance requires more moves to optimally reach the goal. Nowadays, metaheuristic optimization algorithms are widely used for solving complex problems [33], [34], [35]. One of the algorithms that have been recently used to obtain non-optimal solutions to the Fifteen Puzzle problems was a metaheuristic algorithm Artificial Bee Colony (ABC) [36]. Here, the BA* algorithm with HH is compared with the ABC algorithm to show that the obtained results of BA* are sufficiently accurate and much nearer to the optimal results. To increase the effectiveness and performance of the heuristic function of the ABC algorithm, three heuristics PDB, MD, and LC were combined. The ABC algorithm was run on 25 randomly generated solvable instances of the Fifteen Puzzle but the algorithm did not produce an optimal solution for any of them and it provided solutions that are far from the optimum [36]. Tuncer [36] argued that the results produced by the ABC algorithm are acceptable even though the solution lengths are far from the optimal solution lengths. Furthermore, the difference between the solution costs obtained by the ABC algorithm and the optimal solutions for most of the puzzle instances increase when the puzzle instances require more moves to optimally reach the goal. For example, according to Table 3, the solution cost obtained by the ABC algorithm for the first nine puzzle instances that need fewer steps to optimally reach the goal is near to the optima while the rest of the puzzle instances are very far from the optima this is because those instances need more steps to optimally reach the goal. According to this example, the difference between the number of moves obtained by the ABC algorithm and the optimal solution will be big, especially for those states that require 80 moves to reach the goal. On the other hand, an important point about our implementation of the BA* algorithm is that the solution lengths for almost all the Fifteen Puzzle instances are 0 to 6 moves away from the optimal solution lengths even for the difficult states as is shown in Table 3, Table 4 and Table 6.

The BA* algorithm with HH was run on the same 25 initial states and the results obtained by the BA* algorithm are very near to the optimal solutions compared to the results obtained by the ABC algorithm. For example, Table 3 shows that the average number of moves in the solutions which is obtained using the ABC algorithm is 58.76 while the average number of moves in the solutions that are obtained by the BA* algorithm is 50.4. In addition, the average number of moves in the solutions found by BA* is only 1.92 away from the average cost of the optimum solution which is 48.48, while the average number of moves in the solutions found by ABC is 10.28 away from the average cost of the optimum solution. Figure 6 illustrates the obtained results of 25 states presented in Table 3 by the ABC and BA* algorithm.

**Table 3. Comparison of results between BA* algorithm and ABC algorithm**

| NO | INITIAL STATE | Optimal LEN | LEN(ABC) BEST | LEN (BA*) |
|---|---|---|---|---|
| 1 | 1 5 2 7 10 14 11 6 15 12 9 3 13 0 8 4 | 34 | 37 | 34 |
| 2 | 5 6 10 7 1 3 11 8 13 4 15 9 14 0 2 12 | 38 | 43 | 38 |
| 3 | 1 11 6 2 10 13 15 5 3 12 0 4 9 7 14 8 | 40 | 46 | 42 |
| 4 | 6 5 2 7 13 0 10 12 4 1 3 14 9 11 15 8 | 44 | 49 | 46 |
| 5 | 4 3 10 7 6 0 1 2 12 15 5 14 9 13 8 11 | 44 | 52 | 46 |
| 6 | 4 10 3 2 1 0 7 8 9 6 13 15 14 12 11 5 | 44 | 51 | 52 |
| 7 | 3 4 11 2 9 1 14 15 7 6 0 8 5 13 12 10 | 44 | 51 | 44 |
| 8 | 3 10 2 5 15 6 13 4 0 11 1 7 9 12 8 14 | 46 | 52 | 48 |
| 9 | 9 4 0 3 14 7 5 12 15 2 13 6 10 1 8 11 | 46 | 54 | 48 |
| 10 | 7 1 12 10 6 11 15 4 0 2 5 14 3 13 8 9 | 48 | 59 | 50 |
| 11 | 1 13 5 7 14 9 10 12 11 8 2 15 6 0 4 3 | 48 | 62 | 50 |
| 12 | 13 9 5 12 10 2 4 11 3 8 0 7 1 14 6 15 | 48 | 64 | 50 |
| 13 | 2 13 6 1 14 5 11 0 12 4 8 10 9 3 15 7 | 50 | 66 | 50 |
| 14 | 11 3 12 9 2 8 10 14 0 7 15 13 1 6 5 4 | 50 | 68 | 52 |
| 15 | 7 6 15 12 14 1 13 3 0 9 8 4 2 11 5 10 | 50 | 68 | 52 |
| 16 | 5 8 13 15 14 0 1 7 4 6 10 2 11 9 12 3 | 52 | 59 | 56 |
| 17 | 12 2 5 11 10 0 1 6 3 14 8 9 7 4 13 15 | 52 | 62 | 52 |
| 18 | 13 3 2 8 12 0 5 1 11 6 9 15 4 14 7 10 | 52 | 63 | 52 |
| 19 | 7 13 1 4 9 12 8 5 15 14 0 6 11 2 3 10 | 52 | 59 | 52 |
| 20 | 8 11 12 10 2 0 15 1 14 6 4 3 7 9 5 13 | 54 | 61 | 58 |
| 21 | 6 8 12 13 7 2 5 14 9 3 1 15 11 0 10 4 | 54 | 65 | 54 |
| 22 | 9 12 2 5 11 1 10 14 0 4 3 8 6 15 7 13 | 54 | 67 | 60 |
| 23 | 10 12 11 7 8 9 14 5 3 13 4 1 6 0 2 15 | 56 | 69 | 56 |
| 24 | 3 10 14 5 1 12 11 8 15 7 9 6 2 0 13 4 | 56 | 71 | 58 |
| 25 | 9 3 12 5 4 14 6 11 8 7 15 13 10 0 2 1 | 56 | 71 | 60 |
| | **Average** | **48.48** | **58.76** | **50.4** |

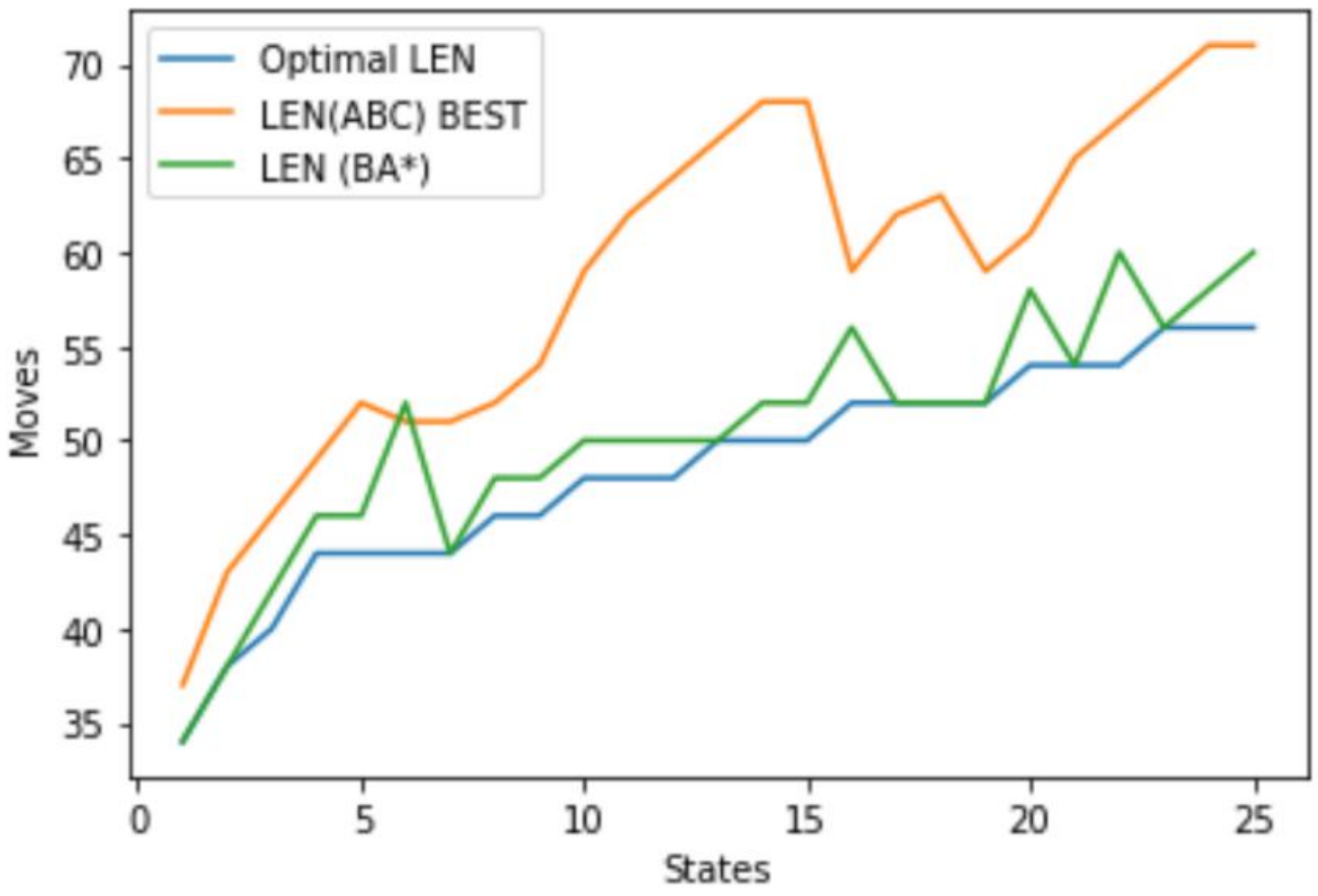

**Figure 6. Results of 25 Fifteen Puzzle states for ABC and BA algorithms**

## 5.2 Bidirectional and Unidirectional Search

In bidirectional search, two separate searches are sequentially or simultaneously run. One search is normal and starts from the initial state toward the goal state, called forward search, and the other search starts from the goal state toward the initial state, called backward search. The search process terminates once when the two searches meet at a common node in the middle and the algorithm constructs a single path that extends from the initial state to the goal state [37][38][39][40]. Pohl [38] was the first one who introduced and implemented a bidirectional heuristic search algorithm with the name Bidirectional Heuristic Path Algorithm (BHPA) and he showed that bidirectional search is more efficient than unidirectional search. BHPA did not work as expected since after the search frontiers meet, both directions of search pass through the opposing frontier to ensure optimality of the solution and this leads to the same node being expanded by the two searches. To resolve this issue, Kwa [41] created a Bidirectional Staged BS* heuristic search algorithm which is derived from Pohl's BHPA algorithm to avoid the re-expansion of a state that has already been expanded in the opposite search. These days, there are several types of research that prove that bidirectional search is very efficient to solve various problems [42][43][44][45][46].

As shown in Algorithm 1, we have implemented a bidirectional search as follows: two sequential processes are run, one branching from the start state, the other branching from the goal state. The first search

“forward search” starts from the initial state and will continue until 75,000 nods are expanded. If the goal state is not reached, the second search “backward search” is initiated from the goal state towards the initial state and this search will continue until it expands 75,000 nodes. If the goal is not found, the backward search stops, and the forward search is performed again. The search process will continue to cycle until both directions meet or the solution is found. During the search, whenever a state is generated by one of the two searches, the algorithm checks if the state has already been generated by the opposite search and if there is it reconstructs a solution path from the two searches. Korf and Schultze [47] were able to compute the number of unique states at each depth of the Fifteen Puzzle. According to [47], the number of generated nodes at each depth gradually increases from depth 0 to depth 53, then the number of generated nodes at each depth starts to gradually decrease from depth 54 to 80. Based on that, the bidirectional search may not be very effective because the number of generated nodes at depth 53 decreases in both directions and it can be difficult for both searches to meet in the middle. Therefore, one of the problems for bidirectional search is that the two searches may not meet or pass by each other without intersecting, but since the A* algorithm retains all the visited nodes in the memory, this ensures that the two searches meet and frontier intersections can be easily tested [9]. Furthermore, there can be more than one optimal solution or non-optimal solution for the Fifteen Puzzle instances that can help the two searches not pass by each other without intersecting [48]. Additionally, bidirectional search is very useful when the problem has not had many goals.

Our implementation of the BA* search can reduce the number of generated states because we use a priority queue to store the estimated costs of states (nodes) and a state in the entire queue (not at a specific level) with the lowest evaluation function value (heuristic value plus the path cost) is always selected to expand. Therefore, the algorithm visits the states in order of their costs not level by level, and results in speeding up the search. Our implementation of BA* search can find optimal or near-optimal solutions even to the difficult states and with a small fraction of states expanded (and stored) compared to Unidirectional A* (UA*) search. Table 4 shows that bidirectional search is more efficient than unidirectional search concerning generated nodes. In Table 4 we run BA* and UA* searches on 28 different states that require 80 moves. The goal state with a blank tile in the top left corner is used for the first 11 instances but the goal state with a blank tile in the bottom right corner is used for the rest of 17 instances. The first nine instances were presented by [49], instances 10 and 11 were presented by [50], and the last 17 states were found by [51]. According to Table 4, BA* search is more efficient than UA* search in terms of node expansion and for ten of the states, UA* is unable to find a solution path and it runs out of memory before finding a solution. Even though the average solution cost obtained by UA* is less than the average solution cost obtained by BA*, the difference is not significant which is only 1.7. Additionally, the average number of states generated by BA* search is significantly less than those generated by UA* search, even though the number of states generated by UA* search for the 10 states has not been counted due to running out of memory.

**Table 4. Comparison of BA* search and UA* search for the 28 difficult Fifteen Puzzle instances requiring 80 moves**

| NO | INITIAL STATE | Optimal LEN | LEN (UA*) | Generated States (UA*) | LEN (BA*) | Generated States (BA*) |
|---|---|---|---|---|---|---|
| 1 | 15 14 8 12 10 11 9 13 2 6 5 1 3 7 4 0 | 80 | Memory runs out | | 88 | 190,569 |
| 2 | 15 11 13 12 14 10 8 9 7 2 5 1 3 6 4 0 | 80 | 84 | 535,435 | 84 | 207,764 |
| 3 | 15 11 13 12 14 10 8 9 2 6 5 1 3 7 4 0 | 80 | 84 | 945,050 | 86 | 373,364 |
| 4 | 15 11 9 12 14 10 13 8 6 7 5 1 3 2 4 0 | 80 | 84 | 1,203,916 | 84 | 420,565 |
| 5 | 15 11 9 12 14 10 13 8 2 6 5 1 3 7 4 0 | 80 | 84 | 1,012,455 | 86 | 203,161 |
| 6 | 15 11 8 12 14 10 13 9 2 7 5 1 3 6 4 0 | 80 | 82 | 2,465,388 | 82 | 186,575 |
| 7 | 15 11 9 12 14 10 8 13 6 2 5 1 3 7 4 0 | 80 | Memory runs out | | 86 | 221,889 |
| 8 | 15 11 8 12 14 10 9 13 2 6 5 1 3 7 4 0 | 80 | Memory runs out | | 86 | 205,463 |
| 9 | 15 11 8 12 14 10 9 13 2 6 4 5 3 7 1 0 | 80 | Memory runs out | | 84 | 195,890 |
| 10 | 15 14 13 12 10 11 8 9 2 6 5 1 3 7 4 0 | 80 | 86 | 181,735 | 84 | 208,946 |
| 11 | 15 11 13 12 14 10 9 5 2 6 8 1 3 7 4 0 | 80 | Memory runs out | | 86 | 538,361 |
| 12 | 0 12 9 13 15 11 10 14 3 7 2 5 4 8 6 1 | 80 | Memory runs out | | 88 | 186,644 |
| 13 | 0 12 10 13 15 11 14 9 3 7 2 5 4 8 6 1 | 80 | 84 | 2,096,287 | 84 | 207,896 |
| 14 | 0 11 9 13 12 15 10 14 3 7 6 2 4 8 5 1 | 80 | 84 | 949,297 | 84 | 198656 |
| 15 | 0 15 9 13 11 12 10 14 3 7 6 2 4 8 5 1 | 80 | 84 | 734,711 | 84 | 167,455 |
| 16 | 0 12 9 13 15 11 10 14 3 7 6 2 4 8 5 1 | 80 | Memory runs out | | 86 | 256,899 |
| 17 | 0 12 14 13 15 11 9 10 3 7 6 2 4 8 5 1 | 80 | 84 | 917,307 | 86 | 205,555 |
| 18 | 0 12 10 13 15 11 14 9 3 7 6 2 4 8 5 1 | 80 | 82 | 1,623,362 | 86 | 341,405 |
| 19 | 0 12 11 13 15 14 10 9 3 7 6 2 4 8 5 1 | 80 | Memory runs out | | 86 | 520,393 |
| 20 | 0 12 10 13 15 11 9 14 7 3 6 2 4 8 5 1 | 80 | 82 | 764,029 | 82 | 199,908 |
| 21 | 0 12 9 13 15 11 14 10 3 8 6 2 4 7 5 1 | 80 | Memory runs out | | 86 | 213,147 |
| 22 | 0 12 9 13 15 11 10 14 8 3 6 2 4 7 5 1 | 80 | 84 | 998,668 | 86 | 205,473 |
| 23 | 0 12 14 13 15 11 9 10 8 3 6 2 4 7 5 1 | 80 | 84 | 1,372,770 | 86 | 416,315 |
| 24 | 0 12 9 13 15 11 10 14 7 8 6 2 4 3 5 1 | 80 | 82 | 1,205,808 | 86 | 213,283 |
| 25 | 0 12 10 13 15 11 14 9 7 8 6 2 4 3 5 1 | 80 | 84 | 105,242 | 84 | 105,242 |
| 26 | 0 12 9 13 15 8 10 14 11 7 6 2 4 3 5 1 | 80 | 82 | 2,259,670 | 86 | 534,581 |
| 27 | 0 12 9 13 15 11 10 14 3 7 5 6 4 8 2 1 | 80 | Memory runs out | | 88 | 160,899 |
| 28 | 0 12 9 13 15 11 10 14 7 8 5 6 4 3 2 1 | 80 | 84 | 2,358,160 | 84 | 209,711 |
| | **Average** | **80** | **83.6** | **1,207,183** | **85.3** | **260,572** |

Table 5 [9] shows the comparison between the two searches UA* and BA* with HH which has been implemented in this paper. According to Table 5, the space and time complexity of the UA* algorithm is $O(b^d)$ where $b$ is the branching and $d$ is the depth of solution, whereas the space and time complexity of BA* algorithm is $O(b^{d/2})$ since in BA* algorithm two searches are run, thus the solution depth is divided by two. One significant point to notice is that the time and space complexity of the A* algorithm strongly depends on the heuristics, which heuristics are used, and how they are implemented [9]. Therefore, in this paper, we took advantage of bidirectional search, the heuristics, and the way of implementing them (as shown in Equation (3)) to reduce the space complexity. Table 5 also presents the completeness and optimality of UA* and BA* with HH. It shows that both the searches are complete but not optimal, this is because of the way of using heuristics as we mentioned before that BA* with HH guarantees either optimal or near-optimal solution.

**Table 5. Evaluation of UA* and BA* searches**

| Criterion | UA* with HH | BA* with HH |
|---|---|---|
| Time Complexity | $O(b^d)$ | $O(b^{d/2})$ |
| Space Complexity | $O(b^d)$ | $O(b^{d/2})$ |
| Complete | Yes | Yes |
| Optimal | **No** | **No** |

## 5.3 Experiments

In the study, the BA* algorithm was applied by combining the advantages of WD, LC, and MD heuristics. The algorithm was run on the 100 random initial states presented by [8], this is mainly to show the efficiency and performance of our implementation of BA*. In Korf's goal state the blank is located at the top left corner, and Korf used IDA* search algorithm with the MD heuristic. Then, those 100 random initial states were used by [14] but this time IDA* algorithm with MD and LC heuristics was run on them, and the result has been added to Table 6. Even though the implementation of the IDA* algorithm with MD and LC heuristics is quite old, we still compare our results with its results because we also use both MD and LC heuristics but in different ways and with another heuristic (WD). Furthermore, Our algorithm extremely reduces the number of generated nodes compared to the results of IDA* with the two heuristics MD and LC.

Table 6 shows that the number of states examined using BA* with HH is much less than the number of states examined using IDA* with MD and LC. For example, the average cost of states examined using IDA* with MD and LC is 37,596,318 states while the average cost of states generated by BA* with HH is only 87,382 states. Furthermore, Table 6 also shows that the average solution cost that is obtained by BA* with HH is about 56.33, and this is very near the average optimal solution cost which is about 53.1 moves. In addition, the number of moves and the number of generated states in the solution of each instance using both algorithms IDA* and BA* are also shown in Table 6. Moreover, it is evident in Table 6 that the solution length of 91% of the instances from 0 to 6 moves far from their optimal solution lengths.

Table 6 also demonstrates the number of state expansions and the WD, MD, and LC values for each of the puzzle instances. Figure 7 shows graphically the total number of states according to the cost difference between their optimal solutions and the solutions achieved by BA* with HH based on Table 6. Table 5 also presents that the solution of 27 states is 0 moves far from optimum (they are optimal solutions), the solution of 22 states is two moves far from optimum, the solution of 24 states is four moves far from optimum, and the solution of 18 states are six moves far from optimum. The figure also shows that only the solution of 6 states is eight moves away from optimum and the solution of three states is 10 moves away from optimum. The last column of Table 6 presents the HH value for each instance and it shows that the HH value is very near to the instance's optimal length. The average HH value, which is 54 is only 0.93 far away from the average optimal length which is 53.07.

Shortly, the most important thing about our implementation of BA* with is that it drastically reduces the search space without consuming a lot of storage space since for all the three heuristics used in this paper only 25KB is required. Furthermore, the results for each instance are very close to the shortest path length even for complex puzzle states. Also, the estimation of HH for each instance shown in Table 6 is near the optimal length.

**Table 6. Comparison of IDA* algorithm with MD and LC and BA* algorithm with HH for the 100 Korf's instances**

| NO | INITIAL STATE | Optimal LEN | IDA* with MD + LC (Generated States) | BA* with HH (Generated States) | LEN (BA*) | BA* with HH (State Expansion) | WD | MD | LC | HH Value |
|---|---|---|---|---|---|---|---|---|---|---|
| 1 | 14 13 15 7 11 12 9 5 6 0 2 1 4 8 10 3 | 57 | 12,205,623 | 183,918 | 67 | 88,032 | 43 | 41 | 0 | 57 |
| 2 | 13 5 4 10 9 12 8 14 2 3 7 1 0 15 11 6 | 55 | 4,556,067 | 111,604 | 59 | 53,919 | 45 | 43 | 2 | 61 |
| 3 | 14 7 8 2 13 11 10 4 9 12 5 0 3 6 1 15 | 59 | 156,590,306 | 47219 | 59 | 22,201 | 43 | 41 | 2 | 59 |
| 4 | 5 12 10 7 15 11 14 0 8 2 1 13 3 4 9 6 | 56 | 9,052,179 | 156,147 | 62 | 75,110 | 44 | 42 | 0 | 58 |
| 5 | 4 7 14 13 10 3 9 12 11 5 6 15 1 2 8 0 | 56 | 2,677,666 | 27,250 | 58 | 12,935 | 44 | 42 | 0 | 58 |
| 6 | 14 7 1 9 12 3 6 15 8 11 2 5 10 0 4 13 | 52 | 4,151,682 | 38478 | 56 | 18,159 | 40 | 36 | 0 | 52 |
| 7 | 2 11 15 5 13 4 6 7 12 8 10 1 9 3 14 0 | 52 | 97,264,710 | 154,743 | 56 | 75,407 | 34 | 30 | 2 | 46 |
| 8 | 12 11 15 3 8 0 4 2 6 13 9 5 14 1 10 7 | 50 | 3,769,804 | 140,971 | 54 | 68,553 | 36 | 32 | 0 | 47 |
| 9 | 3 14 9 11 5 4 8 2 13 12 6 7 10 1 15 0 | 46 | 88,588 | 9539 | 48 | 4,457 | 34 | 32 | 2 | 47 |
| 10 | 13 11 8 9 0 15 7 10 4 3 6 14 5 12 2 1 | 59 | 48,531,591 | 122,553 | 61 | 58,271 | 47 | 43 | 2 | 63 |
| 11 | 5 9 13 14 6 3 7 12 10 8 4 0 15 2 11 1 | 57 | 25,537,948 | 28,847 | 59 | 13,614 | 45 | 43 | 2 | 61 |
| 12 | 14 1 9 6 4 8 12 5 7 2 3 0 10 11 13 15 | 45 | 179,628 | 1335 | 45 | 617 | 37 | 35 | 0 | 49 |
| 13 | 3 6 5 2 10 0 15 14 1 4 13 12 9 8 11 7 | 46 | 1,051,213 | 31505 | 46 | 15,604 | 36 | 36 | 4 | 52 |
| 14 | 7 6 8 1 11 5 14 10 3 4 9 13 15 2 0 12 | 59 | 53,050,799 | 135561 | 63 | 67,256 | 43 | 41 | 4 | 61 |
| 15 | 13 11 4 12 1 8 9 15 6 5 14 2 7 3 10 0 | 62 | 130,071,656 | 117313 | 66 | 55,603 | 46 | 44 | 2 | 63 |
| 16 | 1 3 2 5 10 9 15 6 8 14 13 11 12 4 7 0 | 44 | 2,421,878 | 152,361 | 50 | 75,412 | 24 | 24 | 4 | 36 |
| 17 | 15 14 0 4 11 1 6 13 7 5 8 9 3 2 10 12 | 66 | 100,843,886 | 205,643 | 76 | 98,419 | 54 | 46 | 2 | 71 |
| 18 | 6 0 14 12 1 15 9 10 11 4 7 2 8 3 5 13 | 55 | 5,224,645 | 48027 | 57 | 22,478 | 43 | 43 | 0 | 57 |
| 19 | 7 11 8 3 14 0 6 15 1 4 13 9 5 12 2 10 | 46 | 385,369 | 2279 | 46 | 1,045 | 36 | 36 | 0 | 48 |
| 20 | 6 12 11 3 13 7 9 15 2 14 8 10 4 1 5 0 | 52 | 3,642,638 | 108545 | 54 | 51,873 | 36 | 36 | 0 | 48 |
| 21 | 12 8 14 6 11 4 7 0 5 1 10 15 3 13 9 2 | 54 | 43,980,448 | 157269 | 58 | 75,014 | 40 | 34 | 0 | 51 |
| 22 | 14 3 9 1 15 8 4 5 11 7 10 13 0 2 12 6 | 59 | 79,549,136 | 24166 | 63 | 11,355 | 45 | 41 | 4 | 63 |
| 23 | 10 9 3 11 0 13 2 14 5 6 4 7 8 15 1 12 | 49 | 770,088 | 131754 | 51 | 63,679 | 37 | 33 | 2 | 50 |
| 24 | 7 3 14 13 4 1 10 8 5 12 9 11 2 15 6 0 | 54 | 15,062,608 | 58077 | 54 | 27,678 | 38 | 34 | 0 | 49 |
| 25 | 11 4 2 7 1 0 10 15 6 9 14 8 3 13 5 12 | 52 | 13,453,743 | 6205 | 52 | 2,933 | 36 | 32 | 4 | 51 |
| 26 | 5 7 3 12 15 13 14 8 0 10 9 6 1 4 2 11 | 58 | 50,000,803 | 156,127 | 60 | 75,545 | 42 | 40 | 6 | 61 |
| 27 | 14 1 8 15 2 6 0 3 9 12 10 13 4 7 5 11 | 53 | 31,152,542 | 154,271 | 59 | 76,465 | 37 | 33 | 0 | 48 |
| 28 | 13 14 6 12 4 5 1 0 9 3 10 2 15 11 8 7 | 52 | 1,584,197 | 13084 | 56 | 6,156 | 40 | 36 | 2 | 54 |
| 29 | 9 8 0 2 15 1 4 14 3 10 7 5 11 13 6 12 | 54 | 10,085,238 | 58827 | 58 | 29,134 | 42 | 38 | 2 | 57 |
| 30 | 12 15 2 6 1 14 4 8 5 3 7 0 10 13 9 11 | 47 | 680,254 | 37974 | 47 | 18,219 | 35 | 35 | 0 | 47 |
| 31 | 12 8 15 13 1 0 5 4 6 3 2 11 9 7 14 10 | 50 | 538,886 | 48140 | 56 | 22,949 | 40 | 38 | 2 | 55 |
| 32 | 14 10 9 4 13 6 5 8 2 12 7 0 1 3 11 15 | 59 | 183,341,087 | 169,354 | 65 | 80,244 | 43 | 43 | 6 | 63 |
| 33 | 14 3 5 15 11 6 13 9 0 10 2 12 4 1 7 8 | 60 | 28,644,837 | 179,167 | 68 | 84,823 | 44 | 42 | 2 | 60 |
| 34 | 6 11 7 8 13 2 5 4 1 10 3 9 14 0 12 15 | 52 | 1,174,414 | 2105 | 52 | 984 | 38 | 36 | 12 | 62 |
| 35 | 1 6 12 14 3 2 15 8 4 5 13 9 0 7 11 10 | 55 | 9,214,047 | 41423 | 55 | 19,949 | 41 | 39 | 4 | 58 |
| 36 | 12 6 0 4 7 3 15 1 13 9 8 11 2 14 5 10 | 52 | 4,657,636 | 25466 | 52 | 12,123 | 38 | 36 | 2 | 52 |
| 37 | 8 1 7 12 11 0 10 5 9 15 6 13 14 2 3 4 | 58 | 21,274,607 | 29895 | 58 | 14,354 | 44 | 40 | 4 | 61 |
| 38 | 7 15 8 2 13 6 3 12 11 0 4 10 9 5 1 14 | 53 | 4,946,981 | 19491 | 53 | 9,456 | 41 | 41 | 4 | 59 |
| 39 | 9 0 4 10 1 14 15 3 12 6 5 7 11 13 8 2 | 49 | 3,911,623 | 117816 | 53 | 56,892 | 35 | 35 | 4 | 51 |
| 40 | 11 5 1 14 4 12 10 0 2 7 13 3 9 15 6 8 | 54 | 13,107,557 | 155,875 | 60 | 75,290 | 38 | 36 | 0 | 50 |
| 41 | 8 13 10 9 11 3 15 6 0 1 2 14 12 5 4 7 | 54 | 12,388,516 | 23088 | 54 | 10,979 | 42 | 36 | 0 | 54 |
| 42 | 4 5 7 2 9 14 12 13 0 3 6 11 8 1 15 10 | 42 | 217,288 | 31410 | 42 | 15,122 | 32 | 30 | 0 | 42 |
| 43 | 11 15 14 13 1 9 10 4 3 6 2 12 7 5 8 0 | 64 | 7,034,879 | 24312 | 66 | 11,764 | 54 | 48 | 0 | 70 |
| 44 | 12 9 0 6 8 3 5 14 2 4 11 7 10 1 15 13 | 50 | 3,819,541 | 38881 | 50 | 19,021 | 34 | 32 | 2 | 47 |
| 45 | 3 14 9 7 12 15 0 4 1 8 5 6 11 10 2 13 | 51 | 764,473 | 3920 | 51 | 1,823 | 39 | 39 | 2 | 54 |
| 46 | 8 4 6 1 14 12 2 15 13 10 9 5 3 7 0 11 | 49 | 1,510,387 | 40535 | 49 | 19,597 | 35 | 35 | 2 | 49 |
| 47 | 6 10 1 14 15 8 3 5 13 0 2 7 4 9 11 12 | 47 | 221,531 | 13684 | 47 | 6,591 | 35 | 35 | 0 | 47 |
| 48 | 8 11 4 6 7 3 10 9 2 12 15 13 0 1 5 14 | 49 | 255,047 | 3039 | 49 | 1,431 | 41 | 39 | 0 | 54 |
| 49 | 10 0 2 4 5 1 6 12 11 13 9 7 15 3 14 8 | 59 | 203,873,877 | 159,862 | 63 | 76,448 | 39 | 33 | 6 | 56 |
| 50 | 12 5 13 11 2 10 0 9 7 8 4 3 14 6 15 1 | 53 | 6,225,180 | 90984 | 57 | 43,105 | 41 | 39 | 2 | 56 |
| 51 | 10 2 8 4 15 0 1 14 11 13 3 6 9 7 5 12 | 56 | 4,683,054 | 20341 | 56 | 9,886 | 44 | 44 | 0 | 59 |
| 52 | 10 8 0 12 3 7 6 2 1 14 4 11 15 13 9 5 | 56 | 33,691,153 | 145435 | 60 | 71,856 | 40 | 38 | 4 | 57 |
| 53 | 14 9 12 13 15 4 8 10 0 2 1 7 3 11 5 6 | 64 | 125,641,730 | 194,213 | 70 | 92,613 | 54 | 50 | 0 | 71 |
| 54 | 12 11 0 8 10 2 13 15 5 4 7 3 6 9 14 1 | 56 | 26,080,659 | 140776 | 58 | 66,766 | 42 | 40 | 0 | 55 |
| 55 | 13 8 14 3 9 1 0 7 15 5 4 10 12 2 6 11 | 41 | 163,077 | 7430 | 45 | 3502 | 33 | 29 | 2 | 45 |
| 56 | 3 15 2 5 11 6 4 7 12 9 1 0 13 14 10 8 | 55 | 166,183,825 | 158,326 | 57 | 76,757 | 35 | 29 | 4 | 49 |
| 57 | 5 11 6 9 4 13 12 0 8 2 15 10 1 7 3 14 | 50 | 3,977,809 | 69754 | 52 | 33,060 | 36 | 36 | 4 | 52 |
| 58 | 5 0 15 8 4 6 1 14 10 11 3 9 7 12 2 13 | 51 | 3,563,941 | 36620 | 51 | 17,276 | 39 | 37 | 6 | 57 |
| 59 | 15 14 6 7 10 1 0 11 12 8 4 9 2 5 13 3 | 57 | 90,973,287 | 163,975 | 63 | 76,328 | 39 | 35 | 0 | 51 |
| 60 | 11 14 13 1 2 3 12 4 15 7 9 5 10 6 8 0 | 66 | 256,537,528 | 167,846 | 72 | 80,902 | 54 | 48 | 2 | 72 |
| 61 | 6 13 3 2 11 9 5 10 1 7 12 14 8 4 0 15 | 45 | 672,959 | 8748 | 45 | 4,256 | 31 | 31 | 2 | 43 |
| 62 | 4 6 12 0 14 2 9 13 11 8 3 15 7 10 1 5 | 57 | 8,463,998 | 31031 | 61 | 14,819 | 45 | 43 | 4 | 63 |
| 63 | 8 10 9 11 14 1 7 15 13 4 0 12 6 2 5 3 | 56 | 20,999,336 | 155,693 | 60 | 76,216 | 42 | 40 | 2 | 57 |
| 64 | 5 2 14 0 7 8 6 3 11 12 13 15 4 10 9 1 | 51 | 43,522,756 | 162,795 | 59 | 76,668 | 37 | 31 | 2 | 49 |
| 65 | 7 8 3 2 10 12 4 6 11 13 5 15 0 1 9 14 | 47 | 2,444,273 | 32292 | 49 | 15,837 | 33 | 31 | 2 | 45 |
| 66 | 11 6 14 12 3 5 1 15 8 0 10 13 9 7 4 2 | 61 | 394,246,898 | 173,935 | 69 | 83,771 | 43 | 41 | 0 | 57 |
| 67 | 7 1 2 4 8 3 6 11 10 15 0 5 14 12 13 9 | 50 | 47,499,462 | 156,415 | 52 | 76,051 | 30 | 28 | 2 | 41 |
| 68 | 7 3 1 13 12 10 5 2 8 0 6 11 14 15 4 9 | 51 | 6,959,507 | 67469 | 51 | 32,486 | 33 | 31 | 2 | 45 |
| 69 | 6 0 5 15 1 14 4 9 2 13 8 10 11 12 7 3 | 53 | 5,186,587 | 157,429 | 59 | 75,889 | 37 | 37 | 0 | 49 |
| 70 | 15 1 3 12 4 0 6 5 2 8 14 9 13 10 7 11 | 52 | 40,161,673 | 157,065 | 58 | 75,170 | 36 | 30 | 2 | 48 |
| 71 | 5 7 0 11 12 1 9 10 15 6 2 3 8 4 13 14 | 44 | 539,387 | 153,276 | 54 | 75,984 | 30 | 30 | 0 | 40 |
| 72 | 12 15 11 10 4 5 14 0 13 7 1 2 9 8 3 6 | 56 | 55,514,360 | 175,260 | 60 | 84,165 | 42 | 38 | 4 | 59 |
| 73 | 6 14 10 5 15 8 7 1 3 4 2 0 12 9 11 13 | 49 | 1,130,807 | 3981 | 55 | 1,943 | 41 | 37 | 0 | 53 |
| 74 | 14 13 4 11 15 8 6 9 0 7 3 1 2 10 12 5 | 56 | 310,312 | 39239 | 62 | 18,653 | 48 | 46 | 0 | 63 |
| 75 | 14 4 0 10 6 5 1 3 9 2 13 15 12 7 8 11 | 48 | 5,796,660 | 154,153 | 54 | 75,428 | 30 | 30 | 2 | 42 |
| 76 | 15 10 8 3 0 6 9 5 1 14 13 11 7 2 12 4 | 57 | 25,481,596 | 23489 | 61 | 11,013 | 45 | 41 | 8 | 67 |
| 77 | 0 13 2 4 12 14 6 9 15 1 10 3 11 5 8 7 | 54 | 5,479,397 | 147297 | 56 | 71,101 | 42 | 34 | 0 | 53 |
| 78 | 3 14 13 6 4 15 8 9 5 12 10 0 2 7 1 11 | 53 | 2,722,095 | 14561 | 55 | 6,818 | 43 | 41 | 0 | 57 |
| 79 | 0 1 9 7 11 13 5 3 14 12 4 2 8 6 10 15 | 42 | 107,088 | 13337 | 42 | 6,382 | 30 | 28 | 0 | 39 |
| 80 | 11 0 15 8 13 12 3 5 10 1 4 6 14 9 7 2 | 57 | 39,801,475 | 18182 | 61 | 8,575 | 45 | 43 | 2 | 61 |
| 81 | 13 0 9 12 11 6 3 5 15 8 1 10 4 14 2 7 | 53 | 1,088,123 | 7067 | 53 | 3,315 | 41 | 39 | 2 | 56 |
| 82 | 14 10 2 1 13 9 8 11 7 3 6 12 15 5 4 0 | 62 | 203,606,265 | 175,438 | 68 | 87,036 | 44 | 40 | 2 | 59 |
| 83 | 12 3 9 1 4 5 10 2 6 11 15 0 14 7 13 8 | 49 | 2,155,880 | 81099 | 51 | 40,246 | 35 | 31 | 4 | 49 |
| 84 | 15 8 10 7 0 12 14 1 5 9 6 3 13 11 4 2 | 55 | 17,323,672 | 155,602 | 61 | 76,297 | 39 | 37 | 0 | 51 |
| 85 | 4 7 13 10 1 2 9 6 12 8 14 5 3 0 11 15 | 44 | 933,953 | 15192 | 46 | 7,233 | 32 | 32 | 0 | 43 |
| 86 | 6 0 5 10 11 12 9 2 1 7 4 3 14 8 13 15 | 45 | 237,466 | 9557 | 47 | 4,558 | 35 | 35 | 2 | 49 |
| 87 | 9 5 11 10 13 0 2 1 8 6 14 12 4 7 3 15 | 52 | 7,928,514 | 154,476 | 54 | 75,808 | 36 | 34 | 2 | 49 |
| 88 | 15 2 12 11 14 13 9 5 1 3 8 7 0 10 6 4 | 65 | 422,768,851 | 168,797 | 71 | 80,366 | 49 | 43 | 0 | 63 |
| 89 | 11 1 7 4 10 13 3 8 9 14 0 15 6 5 2 12 | 54 | 29,171,607 | 53615 | 54 | 25,470 | 40 | 38 | 2 | 55 |
| 90 | 5 4 7 1 11 12 14 15 10 13 8 6 2 0 9 3 | 50 | 649,591 | 72881 | 52 | 35,510 | 36 | 36 | 2 | 50 |
| 91 | 9 7 5 2 14 15 12 10 11 3 6 1 8 13 0 4 | 57 | 91,220,187 | 155,268 | 61 | 75,765 | 43 | 41 | 0 | 57 |
| 92 | 3 2 7 9 0 15 12 4 6 11 5 14 8 13 10 1 | 57 | 68,307,452 | 74190 | 57 | 36,186 | 39 | 37 | 2 | 53 |
| 93 | 13 9 14 6 12 8 1 2 3 4 0 7 5 10 11 15 | 46 | 350,208 | 69375 | 50 | 33,288 | 36 | 34 | 2 | 49 |
| 94 | 5 7 11 8 0 14 9 13 10 12 3 15 6 1 4 2 | 53 | 390,368 | 34380 | 59 | 16,093 | 45 | 45 | 2 | 62 |
| 95 | 4 3 6 13 7 15 9 0 10 5 8 11 2 12 1 14 | 50 | 1,517,920 | 69061 | 54 | 32,446 | 42 | 34 | 2 | 55 |
| 96 | 1 7 15 14 2 6 4 9 12 11 13 3 0 8 5 10 | 49 | 1,157,734 | 156,717 | 57 | 76,931 | 37 | 35 | 2 | 51 |
| 97 | 9 14 5 7 8 15 1 2 10 4 13 6 12 0 11 3 | 44 | 166,566 | 43761 | 46 | 21,151 | 32 | 32 | 0 | 43 |
| 98 | 0 11 3 12 5 2 1 9 8 10 14 15 7 4 13 6 | 54 | 41,564,669 | 170,987 | 62 | 82,225 | 38 | 34 | 0 | 49 |
| 99 | 7 15 4 0 10 9 2 5 12 11 13 6 1 3 14 8 | 57 | 18,038,550 | 117970 | 61 | 56,181 | 43 | 39 | 0 | 56 |
| 100 | 11 4 0 8 6 10 5 13 12 7 14 3 1 2 9 15 | 54 | 17,778,222 | 162,353 | 62 | 76,666 | 40 | 38 | 4 | 57 |
| **SUM** | | 5,307 | 3,759,631,814 | 8,738,188 | 5,633 | 4,211,030 | 3,957 | 3,705 | 188 | 5,380 |
| **Average** | | 53.07 | 37,596,318 | 87,382 | 56.33 | 42,110 | 40 | 37 | 2 | 54 |

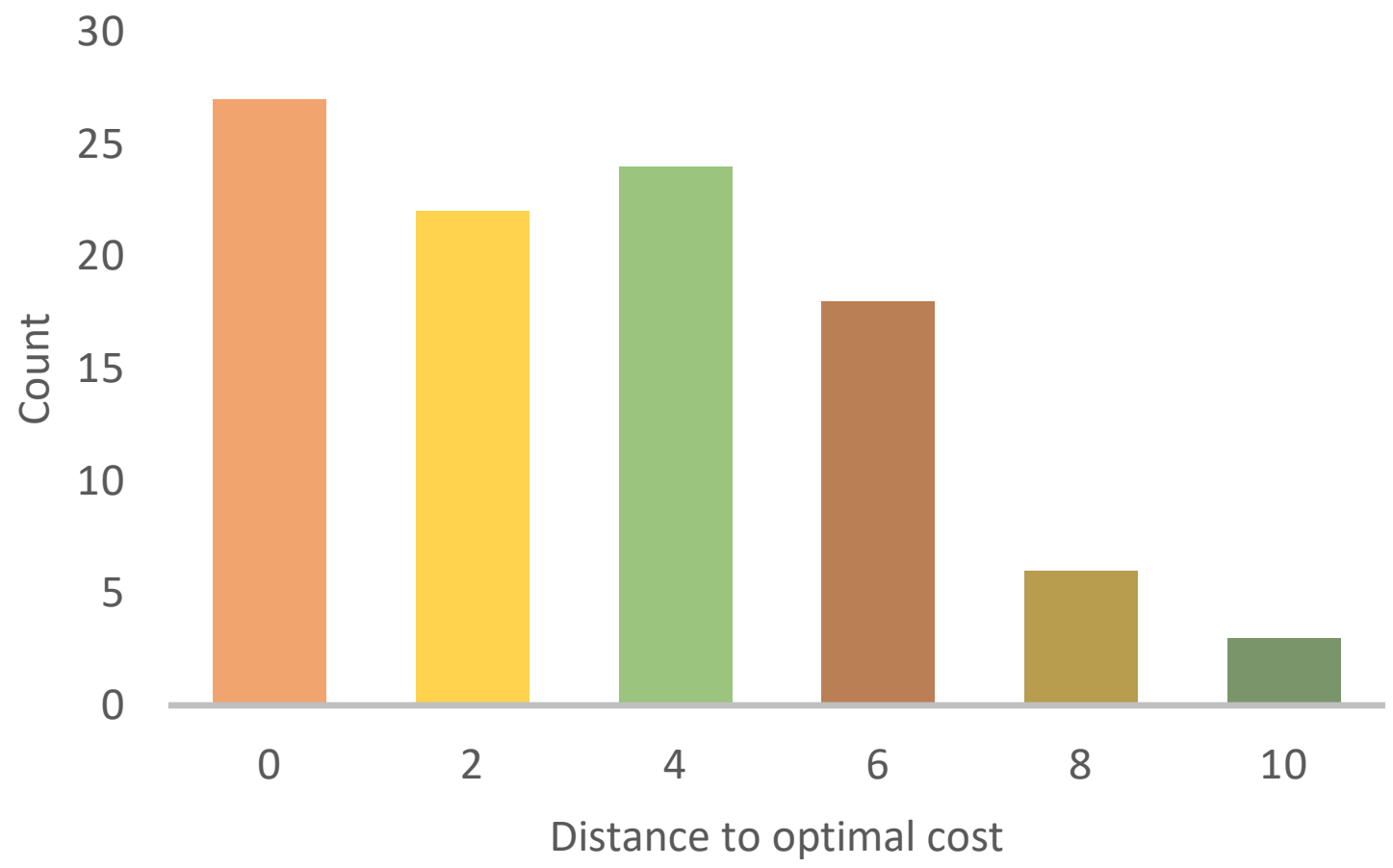

**Figure 7. total number of states according to the cost difference between their optimal solutions and the solutions achieved by BA* with HH**

# 6 Conclusions

In this paper, we proposed Bidirectional A* (BA*) search algorithm with three heuristics WD, LC, and MD, and the heuristics are combined in a way that guides the algorithm efficiently toward the solution and expands fewer states. It is obvious that our implementation of the BA* algorithm does not find the optimal solution for most of the Fifteen Puzzle problem instances but the solutions are very close to optimal length. Additionally, we proved using some empirical evidence that BA* heuristic search algorithm is more efficient than the UA* heuristic search algorithm in terms of state expansions.

Accordingly, designing a heuristic function to accurately choose the next state while exploring the space is challenging due to the huge size of the Fifteen puzzle which is $10^{13}$. To evaluate the performance and efficiency of HH with BA* algorithm, we made some comparisons, especially in terms of optimality and space complexity. We showed that HH with the BA* algorithm produces acceptable results and hugely reduces the search space.

In future work, Hybridizing Heuristic (HH) should be used to increase the effectiveness of metaheuristic algorithms in solving the Fifteen puzzle since HH requires a very small amount of space and it is effective for estimating the complexity of puzzle problems. Therefore, we recommend using novel metaheuristic algorithms such as FDO [52], LPB [53], and ANA [54] for the fifteen puzzles instead of the ABC algorithm since those metaheuristic algorithms work toward optimality.

# Declarations

**Conflict of interest**:
The authors declare no conflict of interest to any party.
The authors have no relevant financial or non-financial interests to disclose.
The authors have no competing interests to declare that are relevant to the content of this article.
All authors certify that they have no affiliations with or involvement in any organization or entity with any financial interest or non-financial interest in the subject matter or materials discussed in this manuscript.
The authors have no financial or proprietary interests in any material discussed in this article.

**Ethical Approval:**
The manuscript is conducted within the ethical manner advised by the targeted journal.

**Consent to Participate**:
Not applicable

**Consent to Publish:** The research is scientifically consented to be published.

**Funding:**
The research received no funds.

**Competing Interests:**
The authors declare no conflict of interest.

**Availability of data and materials:**
Data can be shared upon request from the corresponding author.

**Acknowledgment:** None.